\documentclass[11pt]{article}
\usepackage{acl}
\usepackage{amsmath} 
\usepackage{cleveref}  
\usepackage{times}
\usepackage{latexsym}
\usepackage{makecell}
\usepackage{booktabs}    
\usepackage{multirow}    
\usepackage{array}       
\usepackage[T1]{fontenc}
\usepackage{array}
\newcolumntype{C}[1]{>{\centering\arraybackslash}p{#1}}
\usepackage{subcaption}
\usepackage[utf8]{inputenc}

\usepackage{microtype}
\usepackage[flushleft]{threeparttable} 
\usepackage{inconsolata}
\usepackage{tabularx} 
\usepackage{graphicx}
\usepackage{CJKutf8}

\title{YouZhi: Towards High-Concurrency Financial LLMs \\ via Adaptive GQA-to-MLA Transition}

\author{
  \textbf{Postal Savings Bank of China \& Huawei LLM Team} \\
  Postal Savings Bank of China, Beijing, China \\
  Huawei Technologies, Shenzhen, China
}

\begin{document}
\maketitle

\begin{abstract}
Large language models (LLMs) drive significant financial innovations, yet their high-concurrency deployment is severely bottlenecked by KV cache memory overhead, which inflates infrastructure costs and throttles scalability. To address this, we propose \textbf{YouZhi-LLM}, a highly efficient financial LLM empowered by a comprehensive structural transition and training pipeline natively built on the Huawei Ascend ecosystem. At its algorithmic core, YouZhi-LLM features a \textbf{layer-adaptive GQA-to-MLA transition framework} that dynamically assigns per-layer FreqFold sizes, maximizing KV-cache compression while minimizing perplexity degradation. To recover representation capacity and inject domain expertise, the Ascend-based training pipeline seamlessly integrates generalized knowledge distillation with financial-specific supervised fine-tuning. Evaluations demonstrate the superiority of this systematic approach, with the adaptive transition reducing perplexity degradation by up to 35\% over uniform baselines. Crucially, when evaluated on Ascend NPUs via vLLM-Ascend, the massive KV-cache reduction translates directly into deployment efficiency. Compared to their respective base models, YouZhi-7B yields a 12.3\% improvement in average financial benchmark score alongside a \textbf{2.69$\times$} increase in maximum concurrency; similarly, YouZhi-14B achieves a 7.0\% accuracy gain and a \textbf{2.43$\times$} concurrency boost, establishing a new paradigm for cost-effective, high-throughput financial inference.
\end{abstract}

\section{Introduction}

The rapid iteration of general-purpose large language models (LLMs) has catalyzed a paradigm shift in digital finance, equipping the industry with unprecedented language understanding and reasoning capabilities~\cite{nie2024survey}. While scaling these foundation models unlocks substantial productivity gains across diverse financial applications~\cite{wang2025alpha,kim2024financial}, their deployment in real-time, high-traffic scenarios introduces formidable system-level challenges.

\begin{figure}[!t]
  \centering
  \includegraphics[width=\columnwidth]{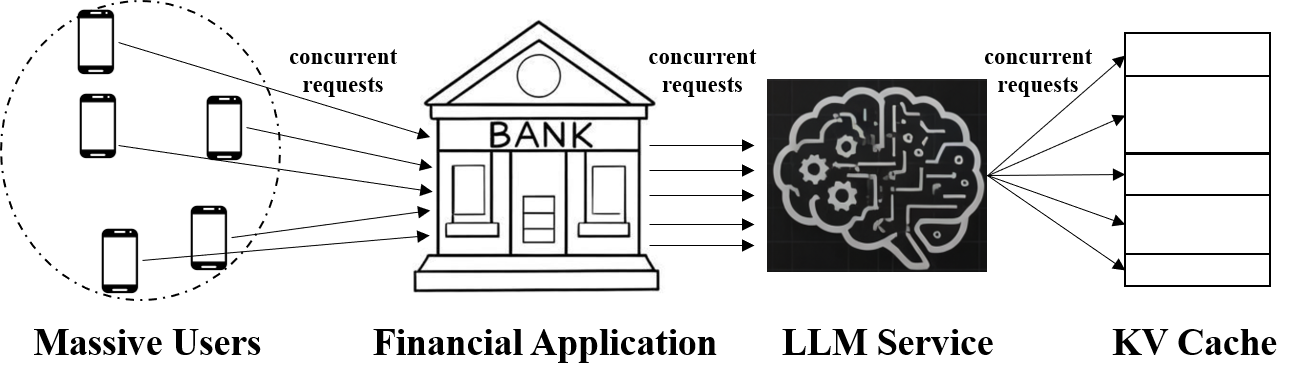}
  \caption{High-Concurrency Mobile Banking.}
  \label{fig:s1}
\end{figure}

Specifically, current advancements in financial LLMs are predominantly gauged by static benchmark accuracy~\cite{xie2024finben}, which obscures a critical gap: the structural disconnect between theoretical performance and production deployability. Real-world financial services, as exemplified by high-concurrency mobile banking scenarios (Figure~\ref{fig:s1}), are governed by stringent operational constraints: \textbf{(i) low latency}, \textbf{(ii) high concurrency}, and \textbf{(iii) high task-completion rates}. Simultaneously, they demand deep, domain-specific expertise that general-purpose models often lack. 

\begin{figure}[!t]
  \centering
  \includegraphics[width=\columnwidth]{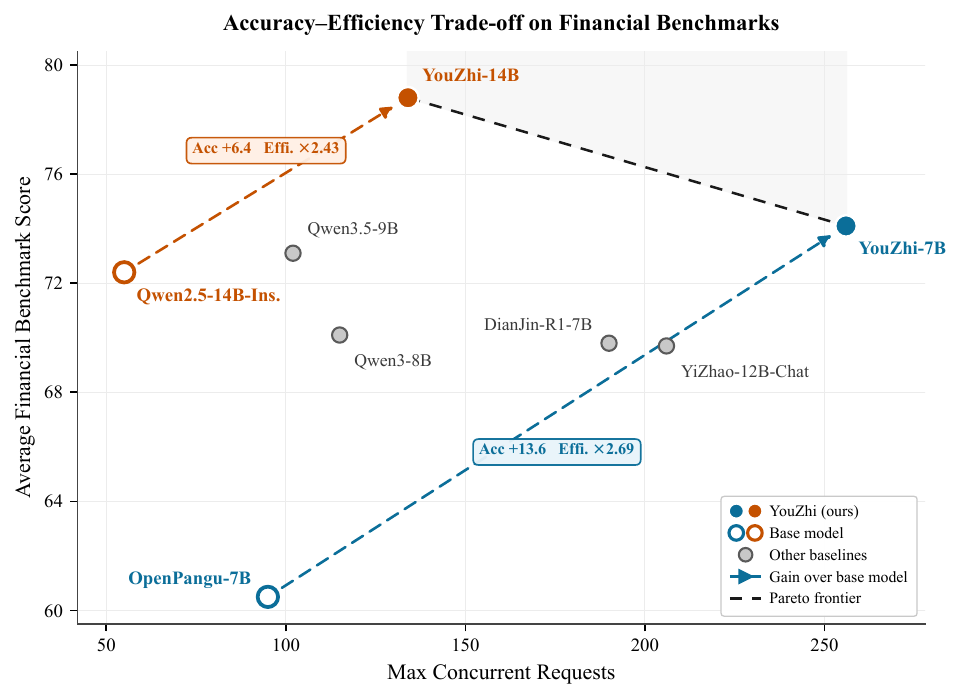}
  \caption{%
    Accuracy--efficiency trade-off on financial benchmarks.
    YouZhi models improve over their base backbones in both average score and
    maximum concurrency, and lie on the Pareto frontier.
  }
  \label{fig:finance-concurrency}
\end{figure}

A pragmatic and economically viable path is to \textbf{harness the robust capabilities of existing foundation models and repurpose them into high-performance financial specialists} through efficient architectural adaptation coupled with comprehensive domain-aware post-training. However, existing financial models—such as BloombergGPT~\cite{wu2023bloomberggpt}, FinGPT~\cite{yang2023fingpt}, YiZhao-12B-Chat~\cite{cmb2024yizhao}, and DianJin-R1~\cite{zhu2025dianjin}—typically inherit the standard Transformer architecture with Grouped-Query Attention (GQA)~\cite{ainslie2023gqa} or Multi-Head Attention (MHA)~\cite{NIPS2017_3f5ee243}, enhanced solely through domain pre-training or supervised fine-tuning. While this paradigm improves financial knowledge understanding~\cite{yang2024financial}, it treats \textit{model capability} and \textit{serving efficiency} as isolated concerns, leaving the fundamental KV cache bottleneck of GQA/MHA architectures unaddressed. As a result, they struggle to achieve an optimal accuracy-efficiency trade-off (as illustrated by the baselines in Figure~\ref{fig:finance-concurrency}).

To bridge this gap, we propose \textbf{YouZhi-LLM}, underpinned by a \textbf{layer-adaptive GQA2MLA transition framework} explicitly tailored for financial deployment. Instead of forcing a one-size-fits-all structural conversion, our approach dynamically optimizes transition parameters (e.g., FreqFold size) on a per-layer basis. This accommodates heterogeneous architectural sensitivities, thereby reaping the KV-cache compression benefits of Multi-Head Latent Attention (MLA) with minimal sacrifice to foundational language modeling capabilities. To recover domain capabilities eroded during the structural transition, we further introduce a \textbf{comprehensive post-training pipeline} featuring stratified data curation and targeted domain augmentation. As demonstrated in Figure~\ref{fig:finance-concurrency}, our resulting models establish a new Pareto frontier, simultaneously maximizing both financial benchmark accuracy and inference concurrency.

Finally, by natively integrating MLA operators on Huawei Ascend clusters, we achieve efficient deployment with substantial memory and latency savings. Our main contributions are summarized as follows:
\begin{itemize}
  \item We conduct a layer-wise analysis of the GQA2MLA transition, revealing that shallow and deep layers exhibit diametrically opposed degradation characteristics during transition.
  \item To capitalize on these layer-specific characteristics, we propose a layer-adaptive transition algorithm that dynamically assigns per-layer FreqFold sizes. Experiments across multiple mainstream LLMs demonstrate a substantial perplexity reduction compared to uniform conversion baselines.
  \item We design a comprehensive post-training pipeline tailored for the converted MLA models, encompassing stratified data compression, financial domain augmentation, refusal data construction for compliance scenarios, and instruction-following reinforcement. This pipeline not only recovers capabilities lost during transition but also yields significant gains on both financial benchmarks and real-world mobile banking tasks.
  \item We implement and deploy the resulting YouZhi-LLM on Huawei Ascend NPUs using the vLLM-Ascend inference framework. Extensive evaluations demonstrate a \textbf{72\% KV cache reduction} and a \textbf{2.69$\times$ improvement in maximum concurrency}, enabling highly efficient, high-concurrency serving for production-grade financial applications.
\end{itemize}

\section{Related Work}
\textbf{Financial Pre-trained Models.} Pre-trained language models have significantly promoted the development of financial natural language processing. Built on the BERT\cite{devlin2019bert} architecture and further pre-trained on massive financial corpora (e.g., financial news, research reports, and corporate filings), FinBERT\cite{yang2020finbert} has become a dominant baseline for financial representation learning. It consistently outperforms vanilla BERT on typical financial tasks, including sentiment analysis, news classification, and named entity recognition.
BBT-FinT5\cite{lu2023bbt} leverages the T5\cite{raffel2020exploring} framework and is adapted to financial summarization, report generation, and sequence prediction. By unifying diverse financial tasks into a generative formulation, BBT-FinT5 delivers strong performance in structured financial text understanding and conditional generation.

With the rapid evolution of large language models (LLMs), research in financial NLP has gradually shifted toward decoder-only GPT-style architectures\cite{brown2020language}, which are inherently better suited for the open-ended generation and causal reasoning required by sophisticated financial scenarios. 
As a representative domain-specific LLM, BloombergGPT\cite{wu2023bloomberggpt} is a 50B-parameter model continually pre-trained on decades of proprietary financial datasets. It yields state-of-the-art results across mainstream financial NLP benchmarks while preserving robust general language proficiency, though its practical application is constrained by its closed-source accessibility. 
In contrast, FinGPT\cite{yang2023fingpt} adopts an open, data-driven development paradigm and integrates LoRA\cite{hu2022lora} fine-tuning and retrieval-augmented generation (RAG) to support low-cost, reproducible domain adaptation, achieving substantial improvements in financial sentiment analysis and market-aware logical reasoning.
Targeting Chinese financial scenarios, YiZhao-12B-Chat\cite{cmb2024yizhao} enhances domain alignment via financial supervised fine-tuning (SFT) and direct preference optimization (DPO)\cite{rafailov2023direct}. Focusing on financial reasoning enhancement, DianJin-R1\cite{zhu2025dianjin} utilizes domain-specific reasoning corpora and reinforcement learning to optimize numerical and compliance reasoning capabilities, outperforming baseline models on complex financial reasoning tasks.

\textbf{Attention Mechanism for Efficient KV Reuse.} To address the KV cache bottleneck, the attention mechanism has been continuously improved. All of these improvements primarily aim to enhance the efficiency of KV reuse. More specifically, the standard MHA \cite{NIPS2017_3f5ee243} maintains completely independent key-value states for each attention head. This introduces significant memory pressure on the KV cache during inference, which becomes the primary bottleneck for batch size and sequence length.
To overcome this, GQA was proposed \cite{ainslie2023gqa}, featuring a grouping mechanism where all attention heads within the same group share a common set of key and value projections. To achieve higher efficiency, MLA  \cite{liu2024deepseek} allows key-value pairs to be jointly compressed across attention heads. This leads to greater reuse efficiency of the KV cache. While other studies such as Sliding Window Attention \cite{yu2026swaaslidingwindowattention} and Linear Attention \cite{NEURIPS2024_d13a3eae} substantially alleviate the memory pressure of KV cache, they sacrifice the ability to capture long-range dependencies, leading to non-negligible performance degradation compared to GQA. As a result, MLA is still a great choice for practical long-context reasoning tasks.

\textbf{GQA to MLA Adaptation} Due to the expensive cost in training MLA model from scratch, it is an efficient way to evolve   existing pre-trained GQA models to MLA structure, such as Transmla \cite{meng2025transmla}, MHA2MLA \cite{ji2025towards}. These studies typically adopt a two-stage process for the transition. In the first stage, positional information of key vector is aggregated into first attention head. This design obviates the need for RoPE in subsequent heads, adapting the structure of partial RoPE used in MLA. In the second stage, the key (with RoPE removed) and value projections are jointly compressed, reducing the KV cache to a lower dimension, typically 512. The perplexity degradation primarily occurs in Stage 1 (partial RoPE) and propagates to Stage 2. Therefore, minimizing the performance drop at partial RoPE Stage constitutes the central challenge in the GQA2MLA process.

\begin{figure}[!t]
  \includegraphics[width=\columnwidth,height=5.5cm]{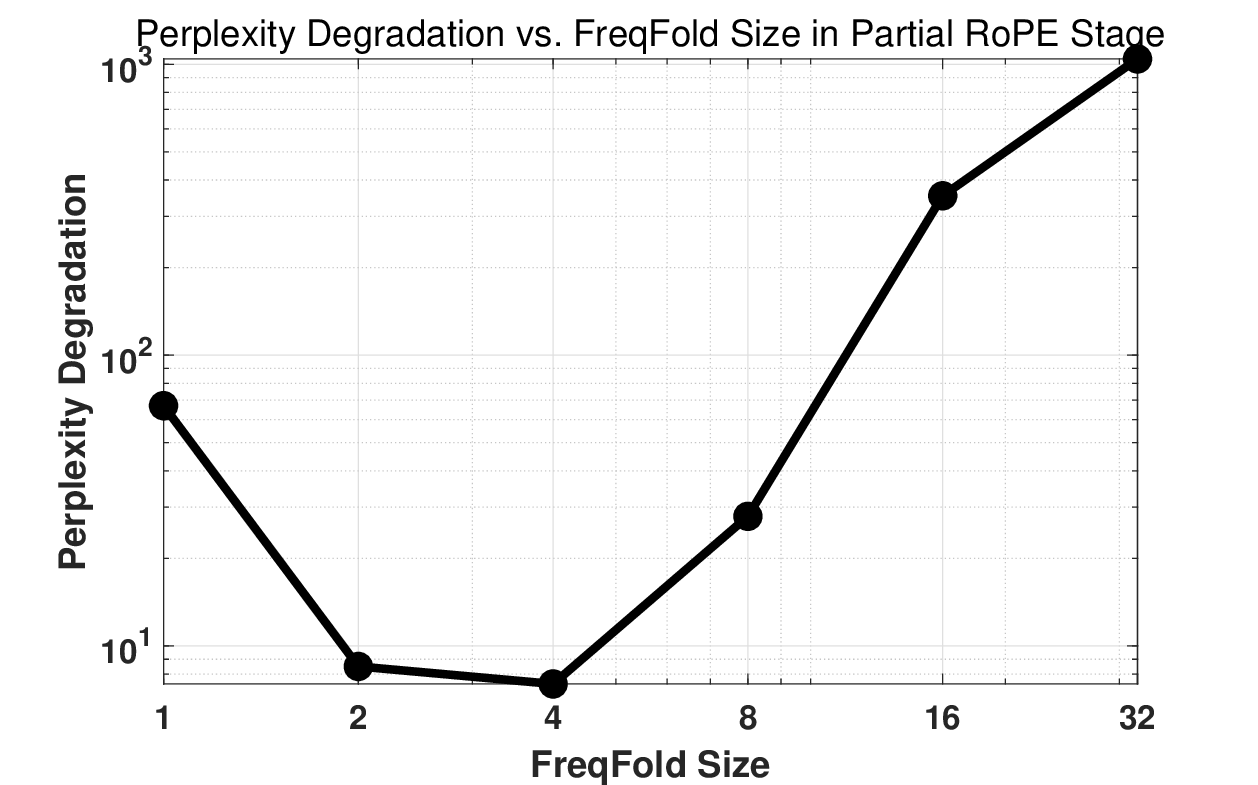}
  \caption{Perplexity degradation of OpenPangu-7B on the WikiText-2 dataset in the partial RoPE modification stage using TransMLA with a RoPE dimension of 128.}
  \label{fig:1}
\end{figure}

In particular, unlike MHA2MLA \cite{ji2025towards} that determines which RoPE dimensions to remove in partial RoPE Stage, TransMLA \cite{meng2025transmla} proposed the FreqFold technique that compresses all RoPE dimensions by grouping adjacent dimensions with similar rotational frequencies for joint PCA. The details of the FreqFold technique are provided in \cite[Appendix C]{meng2025transmla}. Let us denote the number of adjacent dimensions grouped as FreqFold size. While FreqFold technique is theoretically effective, its performance is subject to a critical trade-off: overly aggressive FreqFold size can amplify the approximation error of RoPE frequencies, which may outweigh the benefit of concentrating the principal components. Therefore, careful tuning is required to identify the optimal balance. As shown in Figure~\ref{fig:1}, this trade-off shows that a FreqFold size too large or too small is not the optimal.
What's more, this trade-off manifests differently across transformer layers. Prior methods like TransMLA overlook layer-specific characteristics, applying a uniform FreqFold size to all layers. To address this limitation, we propose a layer-adaptive GQA2MLA transition framework. By leveraging layer-specific characteristics, we dynamically optimizes the FreqFold size for each layer and enhances language modeling capability.

\section{Method}

To investigate the layer-specific characteristics of the GQA2MLA transition, we perform a layer-wise analysis of perplexity degradation in the partial RoPE stage. More specifically in our ablation experiment, only a single layer is translated to MLA at a time, while all other layers remain to be GQA structure unchanged. This allows us to isolate and quantify the layer-specific characteristics in the perplexity degradation of GQA2MLA. 

\begin{figure}[t]
  \includegraphics[width=\columnwidth]{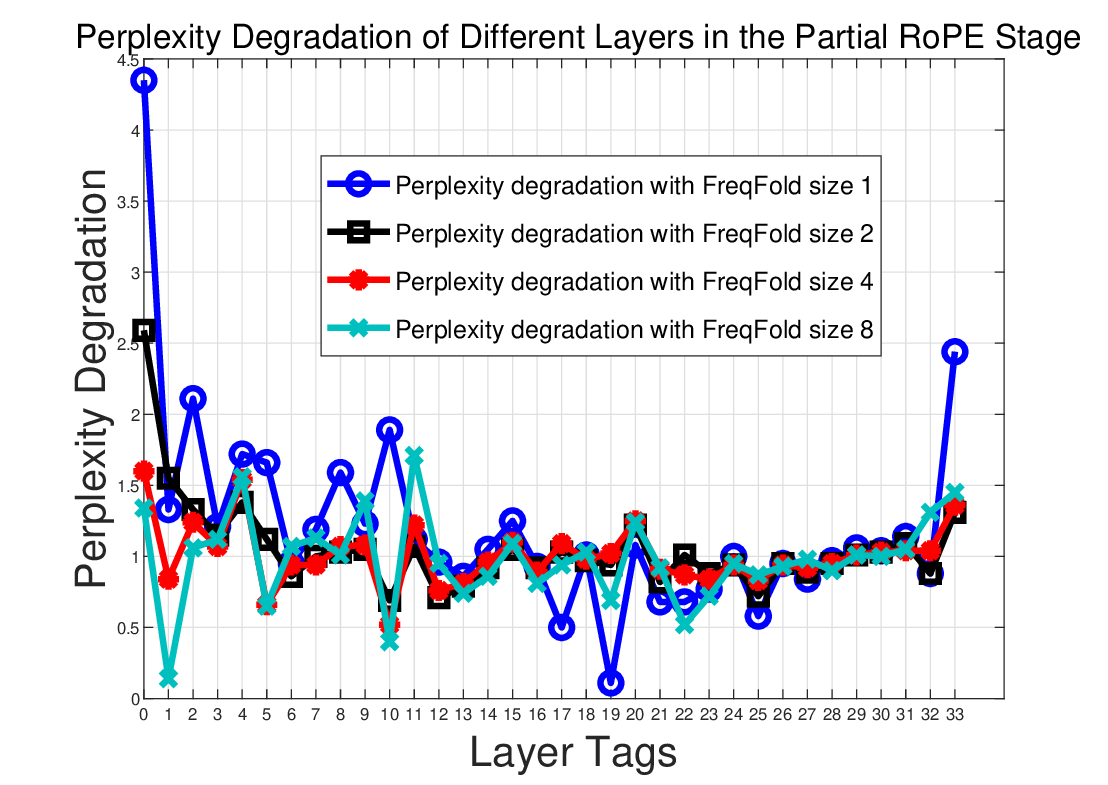}
  \caption{Layer-wise perplexity degradation of OpenPangu-7B on the WikiText-2 dataset in the partial RoPE modification stage using TransMLA with a RoPE dimension of 128.}
  \label{fig:2}
\end{figure}

\noindent\textbf{Observations.} As shown in Figure~\ref{fig:2}, we present the result of layer-wise perplexity degradation in the partial RoPE stage across different FreqFold sizes (e.g., 1, 2, 4, 8). Each polyline corresponds to a fixed FreqFold size. A point on the polyline denotes the perplexity degradation observed when a specific layer is converted to MLA, with all other layers kept in the original GQA structure.
The results reveal a distinct pattern across layers. In the shallow layers (e.g., 0-5), a larger FreqFold size (e.g., 8) yields the lowest perplexity. This suggests that in these layers, the benefit of concentrating principal components outweighs the cost of RoPE frequency approximation. Conversely, in middle layers (e.g., 16-25), the optimal strategy shifts: the minimal perplexity is achieved with a FreqFold size of 1 (effectively no folding). This indicates that in these layers, the frequency approximation error becomes the dominant factor, and preserving precise RoPE frequency information is more critical.
\begin{figure*}[t]
  \begin{subfigure}[b]{0.48\linewidth}
    \includegraphics[width=\linewidth]{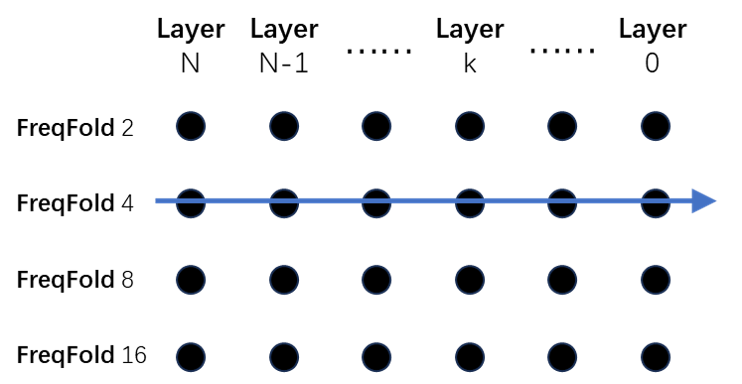}
    \caption{Tansition tajectory diagram of TransMLA.}  
    \label{fig:sub-a}
  \end{subfigure}
  \hfill
  \begin{subfigure}[b]{0.48\linewidth}
    \includegraphics[width=\linewidth]{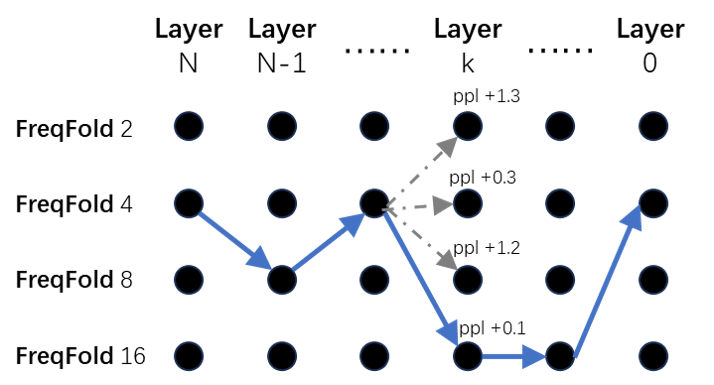}
    \caption{Tansition tajectory diagram of layer-adaptive algorithm.}  
    \label{fig:sub-b}
  \end{subfigure}
  \caption{Schematic Comparison of the Tansition Trajectory: TransMLA vs. Layer-adaptive.}
  \label{fig:3}
\end{figure*}

\noindent\textbf{Insights.} There are significant layer-specific characteristics in the perplexity degradation of GQA2MLA. The FreqFold size should set adaptively for each layer in LLMs.

\subsection{Layer-Adaptive GQA2MLA Transition}
In this section, we introduce a layer-adaptive FreqFold size selection algorithm to minimize perplexity degradation in the GQA2MLA transition. More specifically, we first formulate the layer-adaptive FreqFold selection problem as a combinatorial optimization problem. To tackle the combinatorial explosion of the feasible region with the number of layers, the combinatorial optimization problem is reformulated as a multi-round sequential decision process. Our method optimizes the perplexity in each decision round individually, progressively refining the final solution.

Due to the layer-specific characteristics in the process of GQA2MLA, we should find the optimal FreqFold size for each layer. Let function $d(\alpha_{0}, ..., \alpha_{k}, ..., \alpha_{N})$ denote the perplexity degradation when FreqFold size\footnote{Since the head dimension is typically a power of 2, and the FreqFold size must be a divisor of the head dimension, the feasible choices for FreqFold size are consequently restricted to powers of 2 as well.} of layer $k$ is selected as $\alpha_{k} \in \{0, 2^{0},2^{1},2^{2},..., 2^{h}\}$, where $2^{h}$ is head dimension of model. In particular, when $\alpha_{k}=0$ means that layer $k$ remains GQA structure. 
As a result, the number of feasible solution in the minimization of function $d(\alpha_{0}, ..., \alpha_{k}, ..., \alpha_{N})$ is $(h+1)^{N+1}$. For typical LLMs such as openPangu-7B, the number of feasible solution is $9^{34}$, which is too large to enumerate. 

To overcome this, we should break down this highly complex problem into smaller, manageable subproblems. We can iteratively optimize the FreqFold size for one layer at a time, rather than attempting to solve the intractable global joint optimization directly. To this end, let $C_k$ denote the increase of perplexity degradation attributable to the transition at layer $k$, given that layers after $k$ have been translated. Then $C_k$ can be present as follows,
\begin{equation}
  \label{eq:1}
  C_k= \scalebox{0.82}{$\displaystyle 
     \!\!d(\mathbf{0}_k,\alpha_{k},\dots,\alpha_{N})
     -d(\mathbf{0}_{k+1},\alpha_{k+1},\dots,\alpha_{N})$},
\end{equation}
where $\mathbf{0}_k$ in Eq. (\ref{eq:1}) indicates that the first $k$ layers remain GQA.
Based on the definition of $C_k$, function $d(\alpha_{0}, ..., \alpha_{k}, ..., \alpha_{N})$ can be further decomposed as
\begin{equation}
  \label{eq:2}
  d(\alpha_{0},\dots,\alpha_{k},\dots,\alpha_{N})= \sum_{k=0}^{N} C_k.
\end{equation}
Although we cannot minimize the summation of all terms, i.e. $\sum_{k=0}^{N} C_k$, the optimal solution $\alpha_{k}^{*} $ to minimizing each single term $C_k$ can be easily obtained by simply enumerating the feasible region of $\alpha_{k}$ as follows.
\begin{equation}
\label{eq:3}
\alpha_{k}^{*} = \arg\min_{\alpha_{k}} \{C_k | \alpha_{k+1}, \alpha_{k+2},\dots,\alpha_{N}\}.
\end{equation}
The number of feasible solutions in the optimization problem shown in Eq. (\ref{eq:3}) is only $h+1$. Consequently, the complexity of searching solutions for all layers is $(h+1)\times(N+1)$, which is much smaller than the original complexity $(h+1)^{N+1}$.
\begin{figure*}[t]
  \centering
\includegraphics[width=13cm]{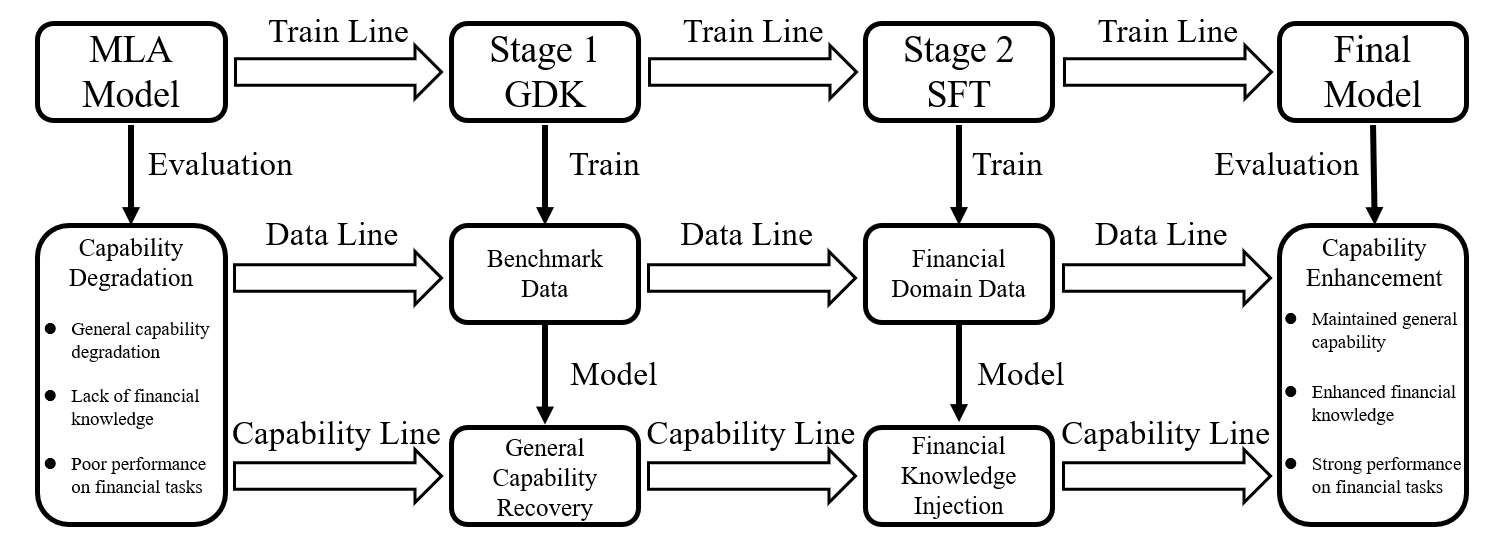}
  \caption{Two-stage Post-training Pipeline.}
  \label{fig:122}
\end{figure*}

To provide an intuitive understanding of the optimization process, we illustrate the transition trajectories of both TransMLA and our method in Figure~\ref{fig:3}. In the schematic, the trajectory of TransMLA is confined to a straight line, as shown in Figure~\ref{fig:sub-a}. This limitation arises because TransMLA employs a uniform FreqFold size across all layers, which inherently restricts its search path. Consequently, the feasible solution space for the GQA2MLA transition is narrowly bounded. To overcome this limitation, we relax the uniform size constraint and propose layer-adaptive strategy given by Eq. (\ref{eq:3}), which allows the independent optimization of the FreqFold size per layer. As shown in Figure~\ref{fig:sub-b}, this flexibility enables our method to find a superior path in the solution space, ultimately achieving lower overall perplexity degradation.

\subsection{Training Pipeline}
In this section, we present a comprehensive post-training pipeline for our translated MLA model, as shown in Figure~\ref{fig:122}. The aim of post-training pipeline is to maintain the model's general capabilities and enhance its performance of downstream tasks in the financial domain.

\begin{figure*}[t]
  \begin{subfigure}[b]{0.48\linewidth}
    \includegraphics[width=\linewidth]{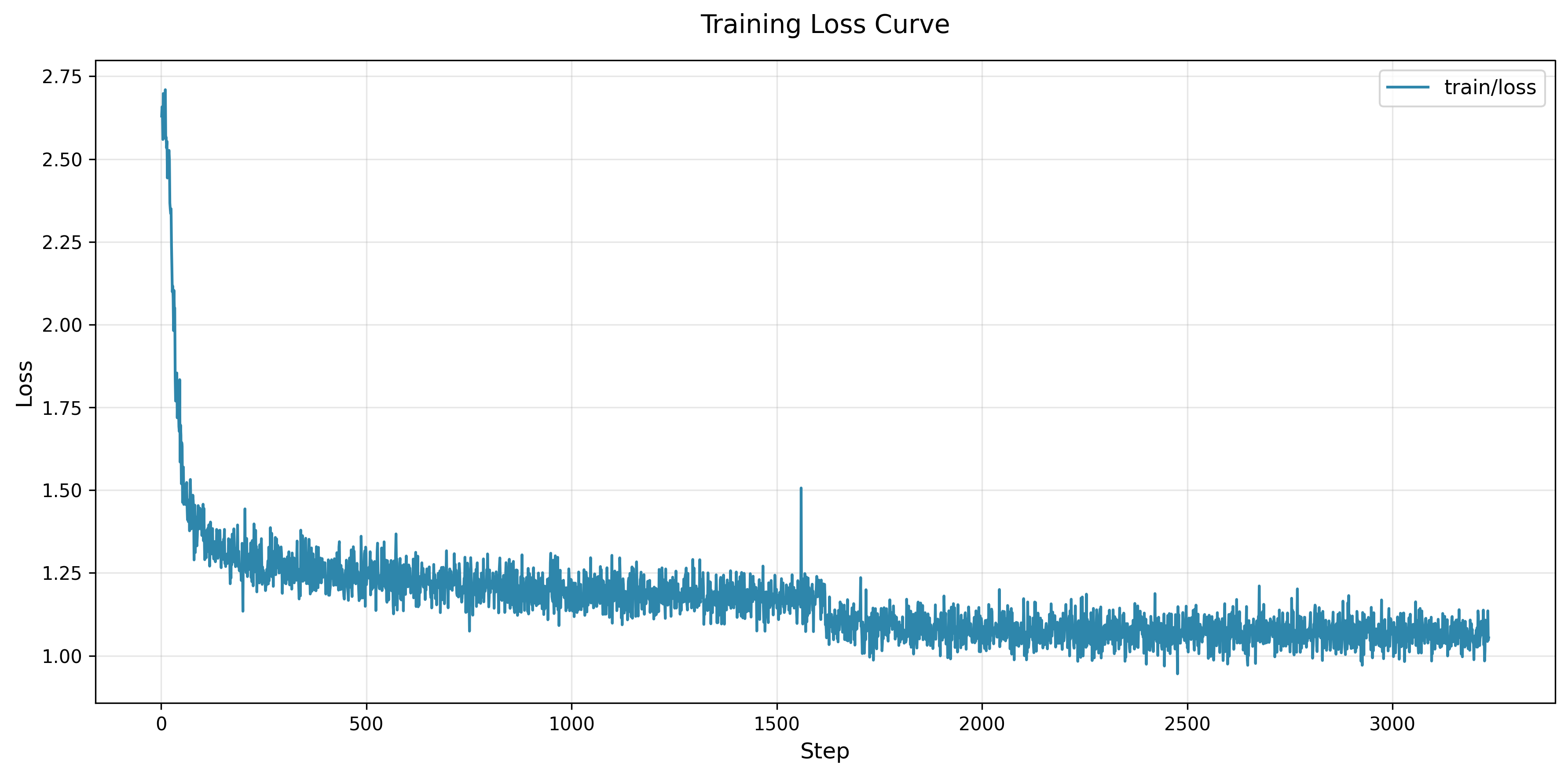}
    \caption{Training loss curve at stage 1: GKD.}  
    \label{fig:dis}
  \end{subfigure}
  \hfill
  \begin{subfigure}[b]{0.48\linewidth}
    \includegraphics[width=\linewidth]{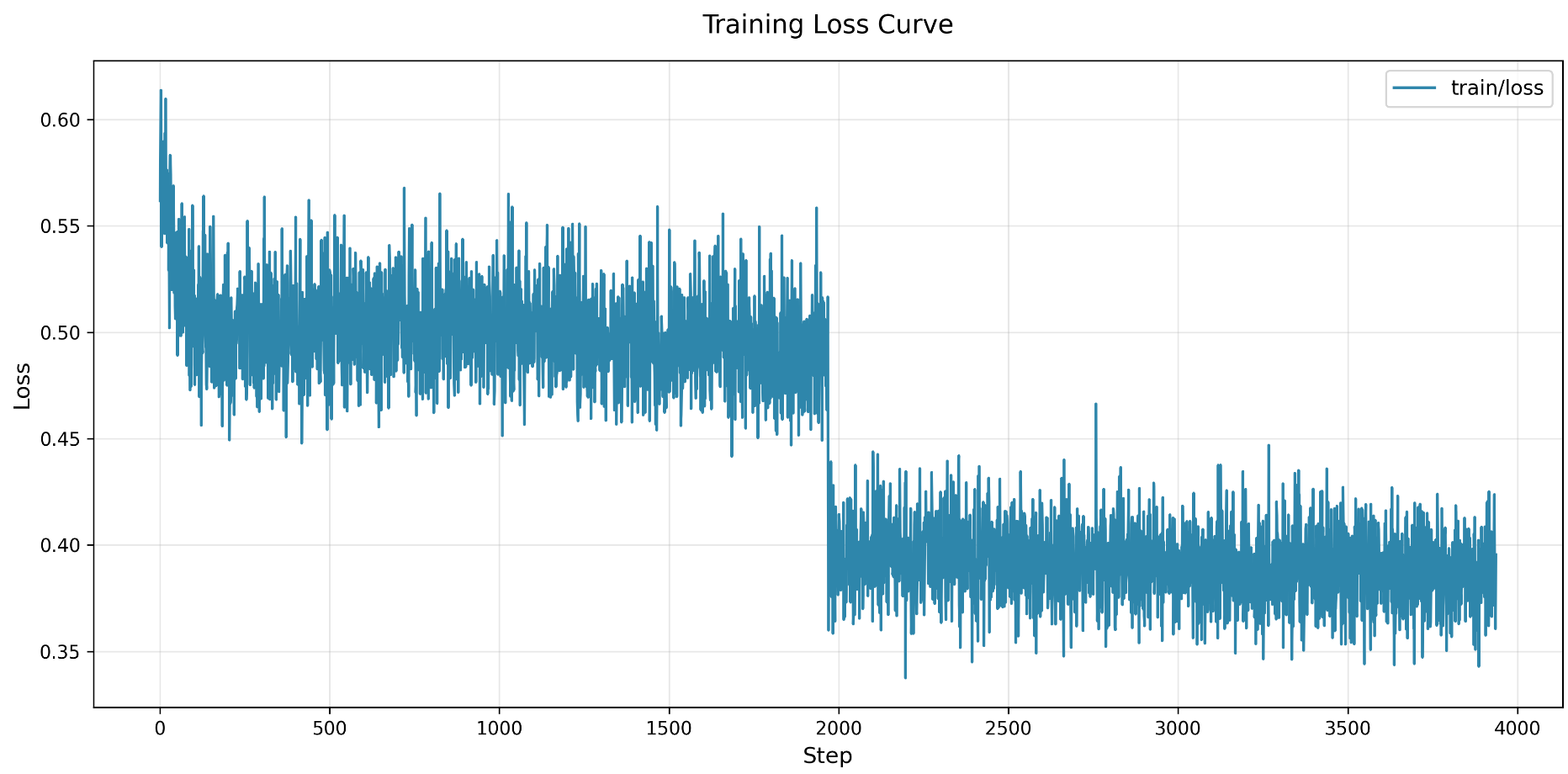}
    \caption{Training loss curve at stage 2: SFT.}  
    \label{fig:sft}
  \end{subfigure}
  \caption{Training loss curve of two-stage post-training pipeline on Ascend A3 Cluster.}
  \label{fig:train}
\end{figure*}

Following the architecture transformation of the base model, we design a two-stage post-training pipeline to develop the domain-specific financial LLM. 
The pipeline was specifically designed to address the unique challenges introduced by GQA2MLA while systematically restoring general-purpose functionalities and improving performance on financial downstream tasks. The entire post-training pipeline is conducted on Ascend clusters, while training loss curve is presented in Figure~\ref{fig:train}.

\noindent\textbf{Stage 1: Generalized Knowledge Distillation.}
The objective of this stage is to restore the general capabilities degraded during the GQA2MLA process.
To this end, we adopt generalized knowledge distillation (GKD) \cite{agarwal2024policy}. 
Specifically, we designate the original GQA-based model as the teacher and the translated MLA model as the student. 
We employ forward KL divergence and an off-policy strategy, training the student on a mixed offline dataset curated from the teacher’s SFT data. This process enables the student to recover performance by learning from the teacher's annotated sequences.

\noindent\textbf{Stage 2: Financial Domain-Specific SFT.}

The objective of this stage is to inject financial knowledge, building on the general capability restored from stage 1. To this end, we construct a comprehensive, multi-source supervised fine-tuning (SFT) dataset that spans all major financial sub-domains, comprising 970,000 high-quality instruction-response pairs. 
The model is fine-tuned on this curated dataset for 2 epochs, optimized with a standard next-token prediction loss. We employ a cosine-decay learning rate schedule, initialized at $1 \times 10^{-5}$
and annealed to 0, coupled with a linear warmup (warmup ratio of 0.01) to stabilize the initial training phase.

\subsection{Construction of Training Data}
The GQA-to-MLA structural transition inevitably introduces accuracy degradation, as discussed in last Section. To recover this degradation and simultaneously enhance the model's financial domain capabilities, carefully constructed training data is essential. In this subsection, we present the data construction pipeline for both supervised fine-tuning (SFT) and reinforcement learning (RL), which aims to restore the general capabilities degraded by the structural transition while improving performance on financial downstream tasks.

\noindent\textbf{Compression of General and Financial Data.}
In the post-training stage, a proper data ratio between general and financial data is critical for simultaneously recovering transition-induced degradation and enhancing financial domain capabilities. However, the available general data typically far exceeds financial data in volume, leading to severe imbalance across capability dimensions. Directly applying entropy-based data compression~\cite{yin2024entropy} to the entire dataset is efficient in that it requires only CPU resources, yet it suffers from poor interpretability and declining effectiveness as data scale increases, since samples of different categories and difficulty levels are compressed together without guaranteeing category completeness, quality priority, or difficulty balance. To address these limitations, we propose a stratified and hierarchical data compression method consisting of three stages. \textit{Stage 1: Tag system construction.} We leverage open-source tagging models to automatically annotate the full dataset across multiple dimensions. For the category dimension, we adopt InsTagger~\cite{lu2024instag} to assign multi-class semantic tags capturing the intent and task type of each sample; for the quality dimension, we employ the DEITA quality scorer~\cite{liu2024deita} to rate each sample on a 1--6 scale; for the difficulty dimension, we use the DEITA complexity scorer to rate sample difficulty on a 1--6 scale. This yields a unified tag system covering category, quality, and difficulty. \textit{Stage 2: Stratified and hierarchical filtering.} Based on the tagging results, the dataset is partitioned into sub-clusters by category tag, and each sub-cluster is further divided by difficulty (easy/medium/hard). Each sub-cluster is assigned a compression target following the priority principle of ``high quality first --- broad domain coverage --- difficulty balance --- high information content'': high-quality samples (above the quality threshold) are retained first, then categories are ensured not to be missing, and finally different difficulty levels are balanced within each category. \textit{Stage 3: Intra-cluster entropy compression.} Within each fine-grained sub-cluster, we apply the entropy-based compression method~\cite{yin2024entropy}. By using lossless compression (zlib) to approximate the incremental information contribution of each candidate sample, we prioritize samples that significantly reduce overall redundancy and remove semantically repetitive or low-information data. Since all data within a cluster share the same category and similar difficulty, the compression process achieves more precise information coverage and avoids the bias caused by cross-category mixed compression.

\noindent\textbf{Augmentation for Missing Domains.}
After the structural transition, domain gaps that were previously tolerable become more pronounced, and open-source data often exhibits significant deficiencies in financial domain coverage, scenarios, and tasks. To systematically fill these gaps, we adopt a progressive diversity data generation approach that expands along the pipeline of ``domain $\rightarrow$ task $\rightarrow$ scenario $\rightarrow$ persona.'' \textit{Diverse persona construction.} Following the meta-decomposition mechanism of DecIF~\cite{hui2025decif}, we first prompt the model to generate meta-domains representing high-level conceptual categories, then produce meta-requests (general task formulations within each domain) and meta-scenarios (concrete situational contexts), and finally combine these elements into specific personas, ensuring systematic coverage and diversity. \textit{Knowledge point extraction and composition.} For missing capability dimensions (e.g., geometry and statistics in mathematics), we use LLMs to extract specific knowledge points from existing relevant data, and then cross-combine these knowledge points to form task descriptions covering different knowledge dimensions. \textit{SFT data construction.} The generated diverse personas and knowledge point compositions are combined to construct corresponding SFT training data. Objective and subjective constraints (e.g., professionalism, compliance, format requirements) are injected during the generation stage, and the same constraint system is applied for automated quality verification after generation, ensuring that the data maintains both diversity and high accuracy.
\begin{figure*}[t]
  \centering
\includegraphics[width=12cm]{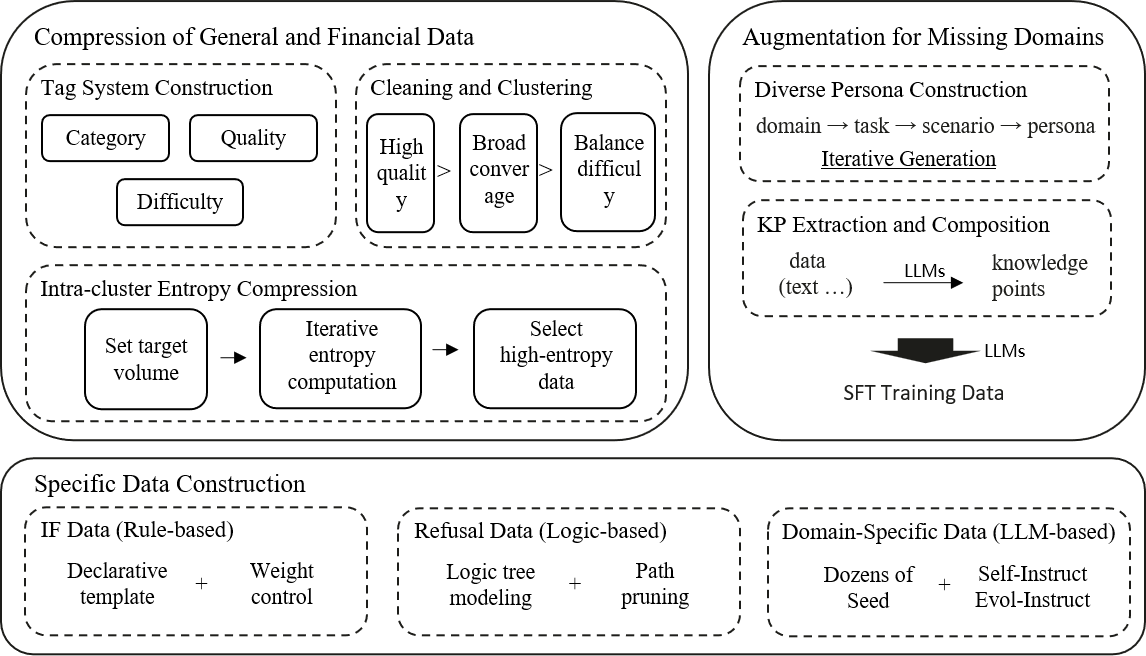}
  \caption{Construction of Training Data.}
  \label{fig:12}
\end{figure*}
\noindent\textbf{Refusal Data Construction.}
Financial scenarios frequently involve incomplete information, such as missing key financial data or contradictory premises, where the model should recognize the insufficiency and respond accordingly. However, since training data typically lacks ``unanswerable'' samples, the model tends to produce hallucinated answers rather than acknowledging ignorance. TreeCut~\cite{ouyang2024treecut} proposes constructing unanswerable questions through logic tree modeling and path pruning, where logical relationships are modeled as dependency trees and information absence is created by selectively removing edges at specific depths. Nevertheless, the original TreeCut method suffers from limited generalization in its rule-based generation mode and unstable quality in its LLM-based generation mode. To address this, we adopt a two-stage strategy: in the first stage, logic tree modeling and path pruning are used to generate a structured framework of unanswerable questions, where controlling the removal depth (\texttt{cutDepth}) precisely constructs information-absent scenarios, ensuring stability and high quality of the underlying information; in the second stage, LLMs translate the tree structures into natural language, converting structured logical relationships into diverse natural language expressions, thereby achieving both diversity and high quality.

\noindent\textbf{Domain-Specific Capability Data Construction.}
The structural transition can cause significant degradation in certain domain-specific capabilities, such as understanding Chinese uppercase financial amounts (e.g., converting \begin{CJK}{UTF8}{min}``壹佰伍拾元整''\end{CJK} to ``150.00'') and performing bidirectional conversions between uppercase and lowercase numerals. To recover these degraded capabilities, we leverage Self-Instruct~\cite{wang2023selfinstruct} and Evol-Instruct~\cite{xu2024wizardlm} to achieve large-scale data amplification from a small number of seed samples. Specifically, we first construct dozens of high-quality seed samples covering typical scenarios and edge cases for the target capability. The Self-Instruct iterative bootstrapping mechanism then guides LLMs to generate new instructions and instances based on the seed samples. Simultaneously, Evol-Instruct progressively evolves simple instructions into more complex variants, expanding the difficulty and coverage of the generated data. Finally, filter validation and similarity-based deduplication ensure the quality and diversity of the generated data. A key advantage of this approach is that it requires only a small number of initial samples (50--200) to achieve large-scale data amplification, making it particularly suitable for domain-specific scenarios where only limited annotated data is available.

\noindent\textbf{Instruction-Following Data Construction.}
After the structural transition, the model often suffers from diminished instruction-following capability, manifesting as an inability to output answers in the required format (e.g., JSON, Markdown, LaTeX) despite producing correct reasoning results. To address this, we select tens of thousands of high-quality domain samples, extract or generate corresponding questions, answers, and reasoning processes, and combine them using predefined instruction templates to batch-construct instruction-following training data. The instruction templates are managed via declarative configuration, where each format constraint type is defined as an independent template node containing the complete format constraint instruction and the standard output format definition. A weight control mechanism flexibly adjusts the sampling ratio of each format type, and a randomization mechanism supports diverse expressions of the same format constraint, thereby establishing the ``instruction-format'' mapping during supervised fine-tuning.
\begin{table}
  \centering
  \small
  \setlength{\tabcolsep}{3pt}
  \begin{tabular}{lcccc}
    \hline
    \textbf{Models} & \textbf{ Orig.} & \textbf{ Trans.} & \textbf{ L-Adap.} &\textbf{ $\Delta$PPL}\\
    \hline
    \verb|OpenPangu-7B|   & 14.1  & 50.3  & 37.6 & -35\%  \\
    \verb|Llama3-8B|   & 6.1  & 25.8  & 12.9 & -65 \% \\
    \verb|Qwen2.5-7B|   & 6.8   & 8.4  & 8.3 & -6 \%  \\
    \verb|Qwen2.5-7B-Ins.|   & 7.5   & 10.0  & 9.5 & -20 \%  \\
    \verb|MiMo-7B-SFT|   & 16.2   & 22.1  & 20.4 & -29\%  \\
    \verb|MiMo-7B-Base|   & 6.9  & 9.6  & 9.0 & -22 \% \\\hline
  \end{tabular}
  \caption{Perplexity on WikiText-2 dataset. "Orig." denotes the original model without any modifications; "Trans."indicates the model modified by TransMLA. "L-Adap." indicates the model modified by our layer-adaptive algorithm. "Ins." denotes instruct model. $\Delta$PPL denotes the reduction of PPL degradation achieved by our algorithm compared to TransMLA. }
  \label{tab:2}
\end{table}
\begin{figure}[t]
  \includegraphics[width=\linewidth]{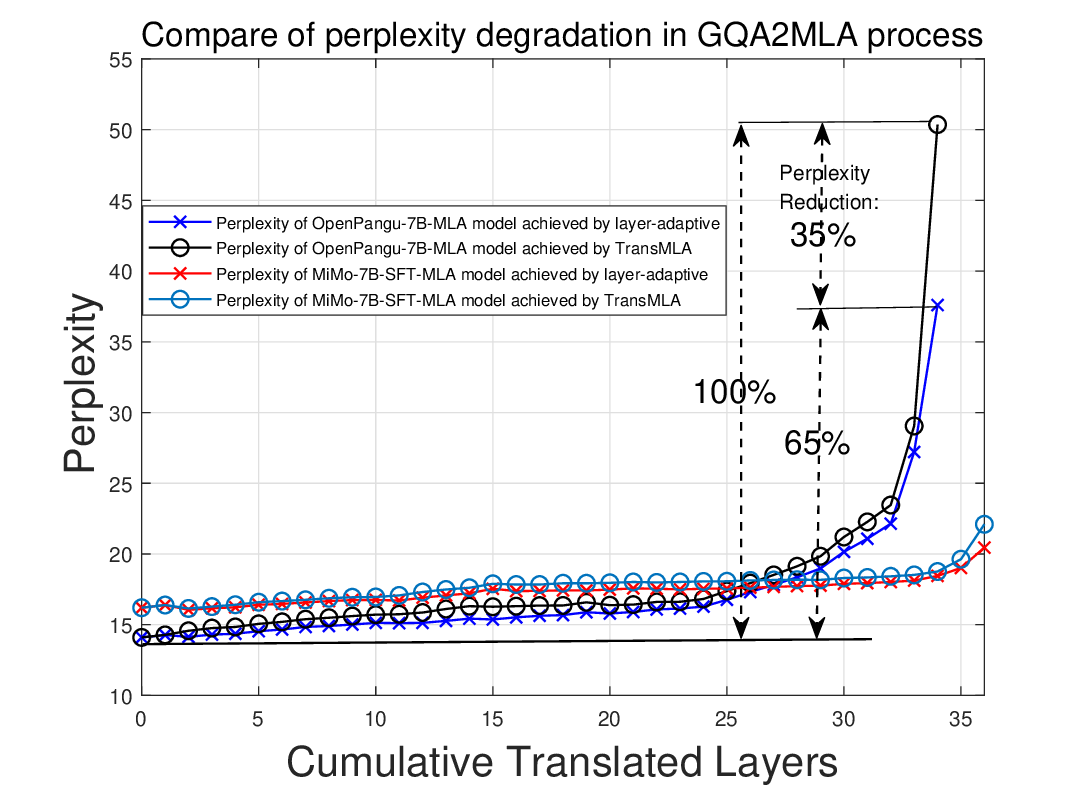}
  \caption{Perplexity Comparison on WikiText-2.}
  \label{fig:5}
\end{figure}

\begin{table*}[!t]
  \centering
  \small
  \begin{tabular}{l l c c c c c c c c}
    \hline
    \textbf{Group} & \textbf{Models} & \textbf{\scalebox{0.88}{C-Eval}} & \textbf{\scalebox{0.88}{IFEval}} & \textbf{\scalebox{0.88} {MATH-500}} & \textbf{\scalebox{0.88}{LCB}} & \textbf{\scalebox{0.88}{H-Swag}} & \textbf{\scalebox{0.88}{SST-5}}  & \textbf{\scalebox{0.88}{CrossNER}} & \textbf{Average}\\
    \hline
    \multirow{2}{*}{Baselines} 
    & YiZhao-12B-Chat    & 70.3    & 43.9  & 46.5  & 3.6   & 70.2  & 54.3  & 49.8 & 48.4 \\
    & DianJin-R1-7B      & 74.5    & 48.8  & 76.8  & 9.3   & 24.4  & 42.9  & 34.2 & 44.4 \\
    \midrule
    \multirow{2}{*}{Base Models}
    & OpenPangu-7B       & 73.4    & 77.8  & 76.6  & 34.6  & 67.8  & 51.5  & 55.2 & 62.4 \\
    & Qwen2.5-14B-Ins.   & 80.2    & 79.6  & 80.6  & 20.4  & 85.9  & 54.3  & 59.7 & 65.8 \\
    \midrule
    \multirow{2}{*}{Our Models}
    & YouZhi-7B          & 72.9    & 77.1  & 76.0  & 30.1  & 90.5  & 62.6  & 58.1 & 66.8 \\
    & YouZhi-14B         & 78.9    & 79.5  & 72.4  & 25.5  & 92.8  & 65.2  & 61.5 & 68.0 \\
    \hline
  \end{tabular}
  \caption{\label{table:general}
   Performance evaluation of general benchmarks.
  }
\end{table*}

\begin{table*}[!t]
  \centering
  \small
  \begin{tabular}{l l C{1.1cm} C{1.1cm} C{1.1cm} C{1.1cm} C{1.1cm} C{1.1cm} C{1.1cm}}
    \hline
    \textbf{Group} & \textbf{Models} & \textbf{\scalebox{0.8}{CFLUE-K}} & \textbf{\scalebox{0.8}{CFLUE-A}} & \textbf{\scalebox{0.8}{FinanceIQ}} & \textbf{\scalebox{0.8}{Fineval}} & \textbf{\scalebox{0.7}{OpenFinData}} & \textbf{\scalebox{0.8}{FPB}} & \textbf{Average} \\
    \hline
    \multirow{4}{*}{Baselines} 
    & YiZhao-12B-Chat    & 62.1   & 39.6   & 63.8   & 84.5   & 86.0   & 82.4   & 69.7  \\
    & DianJin-R1-7B      & 77.6   & 33.7   & 74.7   & 79.0   & 85.8   & 68.2   & 69.8  \\
    & Qwen3-8B           & 66.9   & 37.9   & 69.5   & 80.1   & 83.7   & 82.2   & 70.1  \\
    & Qwen3.5-9B         & 70.7   & 37.1   & 78.1   & 85.7   & 84.7   & 82.4   & 73.1  \\
    \midrule
    \multirow{2}{*}{Base Models}
    & OpenPangu-7B       & 50.7   & 36.5   & 57.2   & 70.1   & 81.7   & 66.5   & 60.5  \\
    & Qwen2.5-14B-Ins.   & 73.4   & 35.6   & 72.7   & 83.3   & 85.6   & 83.5   & 72.4  \\
    \midrule
    \multirow{2}{*}{Our Models}
    & YouZhi-7B          & 71.2   & \textbf{43.3} & 70.2   & 86.5   & 86.7   & 86.8   & 74.1  \\
    & YouZhi-14B         & \textbf{83.9} & 42.4   & \textbf{78.1} & \textbf{91.5} & \textbf{89.4} & \textbf{87.2} & \textbf{78.8} \\  
    \hline
  \end{tabular}
  \caption{\label{table:finance}
Performance evaluation of financial-domain benchmarks.
  }
\end{table*}

\section{Experiments}
We conduct a comprehensive experimental evaluation on Ascend clusters, structured in three aspects: (1) Evaluation on Language Modeling. We benchmark our approach against TransMLA by evaluating perplexity on language modeling dataset WikiText-2  across multiple mainstream LLMs.
(2) Evaluation on Downstream Tasks. We apply the proposed post-training pipeline to the translated MLA model, demonstrating significant accuracy improvements on real-world financial tasks.
(3) Evaluation on High-Concurrency Capability. Leveraging MLA operators on Ascend, we evaluate the system's efficiency, quantifying the reduction in KV cache size and the increase in maximum concurrency on vLLM-Ascend.

\subsection{Experimental Settings}
\textbf{Hardware Platform.} 
All experiments are conducted on Huawei Ascend A3 clusters.

\noindent\textbf{Models and Datasets.} We experiment with official ModelScope checkpoints of GQA model families: Llama \cite{grattafiori2024llama}, Qwen \cite{ahmed2025qwen}, MiMo \cite{xiaomi2025mimo}, and Pangu \cite{chen2025panguembeddedefficientdualsystem}. These GQA models are translated into MLA model with partial RoPE dimension 64 and KV-LoRA rank 512. Specifically, our resulting models, YouZhi-7B and YouZhi-14B, are converted from the GQA models OpenPangu-7B and Qwen2.5-14B-Instruct, respectively.
For evaluation, we report perplexity on WikiText-2 \cite{merity2016pointer} and accuracy on a wide range of benchmarks. These include general-domain benchmarks: C-Eval \cite{huang2023c}, IFEval \cite{zhou2023instruction}, MATH-500 \cite{hendrycks2021measuring}, LiveCodeBench (LCB) \cite{jain2025livecodebench}, HellaSwag (H-Swag) \cite{zellers2019hellaswag}, SST-5 \cite{socher2013recursive}, and CrossNER \cite{liu2021crossner}. We also evaluate on financial-domain benchmarks: CFLUE \cite{zhu2024benchmarking}, FinanceIQ \cite{zhang2023xuanyuan}, FinEval \cite{guo2025fineval}, FBP \cite{malo2014good}, and OpenFinData \cite{openfindata2024}.

\begin{table*}[!t]
  \centering
  \small
  \begin{tabular}{l l C{2.8cm} C{3.8cm} C{2.7cm} }
    \hline
    \textbf{Group} &\textbf{Models} & \textbf{Max Concurrent} & \textbf{Max Throughput} & \textbf{KV Cache Size} \\
    \hline
    \multirow{4}{*}{Baselines} 
    & \verb|DianJin-R1-7B|      & 190  & 5445 tokens/s & 1024 elements \\
    & \verb|YiZhao-12B-Chat|    & 206  & 3692 tokens/s & 512 elements \\
    & \verb|Qwen3-8B|           & 115  & 3442 tokens/s & 2048 elements \\
    & \verb|Qwen3.5-9B|         & 102  & 3040 tokens/s & 2048 elements \\
    \midrule
    \multirow{2}{*}{Base Models} 
    & \verb|OpenPangu-7B|       & 95   & 3325 tokens/s & 2048 elements \\
    & \verb|Qwen2.5-14B-Ins.|  & 55   & 1740 tokens/s & 2048 elements \\
    \midrule
    \multirow{2}{*}{Our Models}
    & \verb|YouZhi-7B|          & 256 (x2.69) & 5865 tokens/s (x1.76) & 576 elements \\
    & \verb|YouZhi-14B|         & 134 (x2.43) & 2990 tokens/s (x1.71) & 576 elements \\
    \hline
  \end{tabular}
  \caption{Performance evaluation of maximum concurrency and throughput in LLM inference service.}
  \label{tab:55}
\end{table*}

\noindent\textbf{Inference Framework.} 
We demonstrate the high-concurrency advantage of our MLA-based model on vLLM-Ascend, a community-maintained branch of vLLM inference framework optimized for Huawei Ascend NPUs.

\subsection{Evaluation on Language Modeling}
\textbf{Overall Comparison.} To validate the language modeling capability enhancement of our algorithm, we conduct experiments on seven representative GQA models (including both base and instruction-tuned variants) and present their perplexity results in Table~\ref{tab:2}. The table compares the performance achieved by our layer-adaptive transition and by the TransMLA baseline. The results show that our layer-adaptive transition consistently achieves lower perplexity than the TransMLA baseline across all tested models. Take Llama3-8B for example, converting the model to MLA via baseline severely degrades perplexity from 6.1 to 25.8. In contrast, our layer-adaptive algorithm results in a much smaller increase, from 6.1 to only 12.9. This represents a 65$\%$ reduction in perplexity degradation compared to baseline.

\noindent\textbf{Ablation Study.} Building on the overall comparison in Table~\ref{tab:2}, we conduct a finer-grained layer-wise ablation study to analyze perplexity. In this experiment, attention layers are progressively translated to MLA, proceeding from the last layer to the first, as visualized in Figure~\ref{fig:5}. The figure plots the perplexity of the resulting hybrid MLA-GQA model against the cumulative number of layers converted. The results demonstrate that, across all conversion steps, the hybrid model generated by our algorithm consistently achieves lower perplexity than the baseline. This consistent superiority confirms that our method is broadly effective and applicable to all attention layers. More importantly, we observe that the most significant performance gap emerges in the first few layers, i.e., the shallowest layers. It indicates that the layer-specific features in these shallow layers are more distinctive, rendering them particularly receptive to and beneficial from our layer-adaptive optimization.

\subsection{Evaluation on Downstream Tasks}
To demonstrate the effectiveness of the post-training pipeline, we conduct a comprehensive evaluation on a wide range of benchmarks (including both general benchmarks and financial-domain benchmarks) and real-world financial applications.

\subsubsection{Evaluation on Benchmarks}

The detailed evaluation results of benchmarks are presented on Table~\ref{table:general} and Table~\ref{table:finance}.
In general benchmarks (Table~\ref{table:general}), YouZhi-7B and YouZhi-14B achieve average scores of 66.8 and 68.0, respectively. This is comparable to or even surpasses their base models, OpenPangu-7B (62.4) and Qwen2.5-14B-Instruct (65.8), demonstrating that our architectural modifications preserve strong general capabilities. More importantly, as shown in financial benchmarks (Table~\ref{table:finance}), our models achieve substantial gains over financial-specific baselines, with YouZhi-7B and YouZhi-14B reaching average scores of 74.1 and 78.8, respectively. This significantly outperforms strong financial baselines such as YiZhao-12B-Chat (69.7) and DianJin-R1-7B (69.8), and even surpasses the more recent general models Qwen3-8B (70.1) and Qwen3.5-9B (73.1). These results highlight the effectiveness of our approach in specializing models for the financial domain without sacrificing general proficiency.
\subsubsection{ Evaluation in Real-World: High Concurrency Financial Applications}
To validate the practical utility of our post-trained model in real-world production, we evaluate it on a representative, high-concurrency financial application scenarios \textit{Mobile Banking}. This scenario encapsulate core interaction demands—precise intent recognition, strict slot filling, and safety-compliant response generation. Both the baseline and our model are finetuned on the same set of proprietary financial data for a fair comparison.

\begin{table}[!t]
\centering
\small
\setlength{\tabcolsep}{5pt}
\begin{tabular}{llcc}
\toprule
\textbf{Task Category} & \textbf{Specific Scenario} & \textbf{Pangu} & \textbf{YouZhi} \\
\hline
\multirow{4}{*}{\makecell[c]{Intent\\Recognition}} 
   & L1 Controller   & 97.8 & 97.5 \\
   & L2 (Loan)  & 97.5 & 97.6 \\
   & L2 (Wealth)  & 99.4 & 99.8 \\
   & L2 (Transfer) & 96.1 & 96.0 \\
\midrule
\multirow{2}{*}{Slot Filling} 
   & Progress Query & 98.1 & 98.1 \\
   & Loan Calculator & 100.0 & 100.0 \\
\bottomrule
\end{tabular}
\caption{Financial Application of Mobile Banking: Pangu (OpenPangu-7B) vs. YouZhi (YouZhi-7B).}
\label{tab:finance-application}
\end{table}

More specifically, we evaluate the models on six critical tasks within the mobile banking pipeline, spanning two levels of intent controllers and two slot-filling scenarios. The L1 controller routes user requests to coarse-grained business domains (Loan, Wealth, Transfer), while L2 controllers perform fine-grained sub-domain classification. Slot-filling tasks require the model to extract structured parameters (e.g., loan amount, transaction time) from informal user utterances. As shown in Table~\ref{tab:finance-application}, YouZhi-7B delivers comparable performance to its base model OpenPangu-7B across all tasks. It achieves an average accuracy of 97.73\% on the four intent recognition tasks and average accuracy of 99.05\% on the two slot-filling tasks. These results confirm model's reliable handling of multi-turn, task-oriented dialogues in mobile banking service.

\subsection{Evaluation on High-Concurrency }
To demonstrate the high-concurrency advantage of the MLA-based model built with our approach, we performed a service pressure test on a single Ascend chip.
The results, summarized in Table~\ref{tab:55}, are obtained from evaluations conducted on the vLLM-Ascend inference framework, with each concurrent request processing 1k input tokens and generating 1k output tokens. As we can see, the maximum concurrency increases from 95 (original OpenPangu-7B) to 256 (our YouZhi-7B ), with the corresponding peak throughput rising from 3,325 tokens/s to 5,865 tokens/s. By significantly reducing the KV cache size from 2,048 to 576 elements, YouZhi-7B achieves a 72\% KV cache reduction and a 2.69× improvement in maximum concurrency.

\section{Conclusion}
In this paper, we present YouZhi-LLM, an efficient financial large language model natively built on the Huawei Ascend ecosystem. To achieve a performant yet deployment-efficient model, we address two key challenges: enhancing the architectural efficiency of the base model and adapting it to complex financial scenarios. First, we introduce a layer-adaptive algorithm that extends the TransMLA framework, effectively capturing layer-specific characteristics and yielding substantial improvements in core language modeling capabilities. Building upon this optimized architecture, we then design a two-stage post-training pipeline that equips the model with robust performance in real-world financial applications such as mobile banking. Finally, by leveraging the vLLM-Ascend inference framework, we demonstrate that YouZhi-LLM not only excels in downstream financial tasks but also supports ultra-high concurrency services on Ascend hardware. The overall workflow establishes an effective pathway for developing and deploying high-performance, industry-specific LLMs within Huawei Ascend ecosystem.



\section{Contributions and Acknowledgments}

This project is a joint effort between the Postal Savings Bank of China (PSBC) and Huawei Technologies. We deeply appreciate the dedication of all members involved. 
\begingroup
\renewcommand\thefootnote{}
\footnotetext{\raggedright $^\star$ Core contributors to this project. \\ $^\dagger$ Correspondence should be addressed to: \texttt{feixiuhong@psbcoa.com.cn}, \texttt{renxiaozhe@huawei.com}.}
\endgroup

\subsection*{Sponsors}
Xinzhuang Niu, Zhaohui Xu, Hang Wang, \\ Yibo He.
\subsection*{Project Leads}
Zhipeng Zhang, Yulong Li, Jing Zhu, Rui Yuan, Jia Yuan, Jiabin Li, Pei Li, Xiuhong Fei, Xiaozhe Ren.

\subsection*{PSBC Team}
Ruihan Long$^\star$, Tianan Zhang$^\star$, Yaozong Wu$^\star$, Chang Liu$^\star$, Wenshuang Yang$^\star$, Zhihao Song$^\star$, Wenjing Xu$^\star$, Shupei Sun$^\star$, Jing Hu, Xinyu Wang, Zequn Ding, Man Luo, Linkai Hou, Hu Zhao, Yang Zhao, Shucheng Lin, Wei Yu, Chenghan Jiang, Jingjing Ding, Jiahui Zhang, Tian Jin, Yuhang Zhang, Xiuhong Fei$^\dagger$, Pei Li, Jiabin Li, Jia Yuan, Rui Yuan, Jing Zhu, Yulong Li, Zhipeng Zhang, Hang Wang, Zhaohui Xu, Xinzhuang Niu.

\subsection*{Huawei Team}
Junjie Wu$^\star$, Duo Zhang$^\star$, Jinbin Fu$^\star$, Zhentao Tang$^\star$, Xin Wang$^\star$, Ning Huang$^\star$, Shuai Zong$^\star$, Sen Wang$^\star$, Bin Wang, Junkui Ju, Jie Ran, Shixiong Kai, Kaichao Liang, Dong Guo, Wei Sun, Jun Xie, Jianwei Li, Lei Cao, Rui Zhao, Mingxuan Yuan, Zhangcheng Lv, Xin Jiang, Xiaozhe Ren$^\dagger$, Yibo He.

\clearpage

\bibliography{custom}

@inproceedings{NIPS2017_3f5ee243,
 author = {Vaswani, Ashish and Shazeer, Noam and Parmar, Niki and Uszkoreit, Jakob and Jones, Llion and Gomez, Aidan N and Kaiser, \L ukasz and Polosukhin, Illia},
 booktitle = {Advances in Neural Information Processing Systems},
 editor = {I. Guyon and U. Von Luxburg and S. Bengio and H. Wallach and R. Fergus and S. Vishwanathan and R. Garnett},
 pages = {},
 publisher = {Curran Associates, Inc.},
 title = {Attention is All you Need},
 url = {https://proceedings.neurips.cc/paper_files/paper/2017/file/3f5ee243547dee91fbd053c1c4a845aa-Paper.pdf},
 volume = {30},
 year = {2017}
}

@article{huang2023c,
  title={C-eval: A multi-level multi-discipline chinese evaluation suite for foundation models},
  author={Huang, Yuzhen and Bai, Yuzhuo and Zhu, Zhihao and Zhang, Junlei and Zhang, Jinghan and Su, Tangjun and Liu, Junteng and Lv, Chuancheng and Zhang, Yikai and Fu, Yao and others},
  journal={Advances in neural information processing systems},
  volume={36},
  pages={62991--63010},
  year={2023}
}

@article{zhou2023instruction,
  title={Instruction-following evaluation for large language models},
  author={Zhou, Jeffrey and Lu, Tianjian and Mishra, Swaroop and Brahma, Siddhartha and Basu, Sujoy and Luan, Yi and Zhou, Denny and Hou, Le},
  journal={arXiv preprint arXiv:2311.07911},
  year={2023}
}

@article{hendrycks2021measuring,
  title={Measuring mathematical problem solving with the math dataset},
  author={Hendrycks, Dan and Burns, Collin and Kadavath, Saurav and Arora, Akul and Basart, Steven and Tang, Eric and Song, Dawn and Steinhardt, Jacob},
  journal={arXiv preprint arXiv:2103.03874},
  year={2021}
}

@inproceedings{jain2025livecodebench,
  title={Livecodebench: Holistic and contamination free evaluation of large language models for code},
  author={Jain, Naman and Gu, Alex and Li, Wen-Ding and Yan, Fanjia and Zhang, Tianjun and Wang, Sida and Solar-Lezama, Armando and Sen, Koushik and Stoica, Ion},
  booktitle={International Conference on Learning Representations},
  volume={2025},
  pages={58791--58831},
  year={2025}
}

@inproceedings{zellers2019hellaswag,
  title={Hellaswag: Can a machine really finish your sentence?},
  author={Zellers, Rowan and Holtzman, Ari and Bisk, Yonatan and Farhadi, Ali and Choi, Yejin},
  booktitle={Proceedings of the 57th annual meeting of the association for computational linguistics},
  pages={4791--4800},
  year={2019}
}

@article{malo2014good,
  title={Good debt or bad debt: Detecting semantic orientations in economic texts},
  author={Malo, Pekka and Sinha, Ankur and Korhonen, Pekka and Wallenius, Jyrki and Takala, Pyry},
  journal={Journal of the Association for Information Science and Technology},
  volume={65},
  number={4},
  pages={782--796},
  year={2014},
  publisher={Wiley Online Library}
}

@inproceedings{zhang2023xuanyuan,
  title={Xuanyuan 2.0: A large chinese financial chat model with hundreds of billions parameters},
  author={Zhang, Xuanyu and Yang, Qing},
  booktitle={Proceedings of the 32nd ACM international conference on information and knowledge management},
  pages={4435--4439},
  year={2023}
}

@misc{openfindata2024,
  title        = {OpenFinData: A Large Language Model Open-source Financial Evaluation Dataset},
  author       = {Shanghai AI Laboratory and East Money},
  howpublished = {\url{https://github.com/open-compass/OpenFinData}},
  year         = {2024},
  note         = {Version 1.0}
}

@inproceedings{guo2025fineval,
  title={Fineval: A chinese financial domain knowledge evaluation benchmark for large language models},
  author={Guo, Xin and Xia, Haotian and Liu, Zhaowei and Cao, Hanyang and Yang, Zhi and Liu, Zhiqiang and Wang, Sizhe and Niu, Jinyi and Wang, Chuqi and Wang, Yanhui and others},
  booktitle={Proceedings of the 2025 Conference of the Nations of the Americas Chapter of the Association for Computational Linguistics: Human Language Technologies (Volume 1: Long Papers)},
  pages={6258--6292},
  year={2025}
}

@inproceedings{zhu2024benchmarking,
  title={Benchmarking large language models on CFLUE-a Chinese financial language understanding evaluation dataset},
  author={Zhu, Jie and Li, Junhui and Wen, Yalong and Guo, Lifan},
  booktitle={Findings of the Association for Computational Linguistics: ACL 2024},
  pages={5673--5693},
  year={2024}
}

@inproceedings{liu2021crossner,
  title={Crossner: Evaluating cross-domain named entity recognition},
  author={Liu, Zihan and Xu, Yan and Yu, Tiezheng and Dai, Wenliang and Ji, Ziwei and Cahyawijaya, Samuel and Madotto, Andrea and Fung, Pascale},
  booktitle={Proceedings of the AAAI conference on artificial intelligence},
  volume={35},
  number={15},
  pages={13452--13460},
  year={2021}
}

@inproceedings{socher2013recursive,
  title={Recursive deep models for semantic compositionality over a sentiment treebank},
  author={Socher, Richard and Perelygin, Alex and Wu, Jean and Chuang, Jason and Manning, Christopher D and Ng, Andrew Y and Potts, Christopher},
  booktitle={Proceedings of the 2013 conference on empirical methods in natural language processing},
  pages={1631--1642},
  year={2013}
}

@article{nie2024survey,
  title={A survey of large language models for financial applications: Progress, prospects and challenges},
  author={Nie, Yuqi and Kong, Yaxuan and Dong, Xiaowen and Mulvey, John M and Poor, H Vincent and Wen, Qingsong and Zohren, Stefan},
  journal={arXiv preprint arXiv:2406.11903},
  year={2024}
}

@article{xie2024finben,
  title={Finben: A holistic financial benchmark for large language models},
  author={Xie, Qianqian and Han, Weiguang and Chen, Zhengyu and Xiang, Ruoyu and Zhang, Xiao and He, Yueru and Xiao, Mengxi and Li, Dong and Dai, Yongfu and Feng, Duanyu and others},
  journal={Advances in Neural Information Processing Systems},
  volume={37},
  pages={95716--95743},
  year={2024}
}

@inproceedings{agarwal2024policy,
  title={On-policy distillation of language models: Learning from self-generated mistakes},
  author={Agarwal, Rishabh and Vieillard, Nino and Zhou, Yongchao and Stanczyk, Piotr and Ramos Garea, Sabela and Geist, Matthieu and Bachem, Olivier},
  booktitle={International Conference on Learning Representations},
  volume={2024},
  pages={21246--21263},
  year={2024}
}

@misc{chen2025panguembeddedefficientdualsystem,
      title={Pangu Embedded: An Efficient Dual-system LLM Reasoner with Metacognition}, 
      author={Hanting Chen and Yasheng Wang and Kai Han and Dong Li and Yunhe Wang et al},
      year={2025},
      eprint={2505.22375},
      archivePrefix={arXiv},
      primaryClass={cs.CL},
      url={https://arxiv.org/abs/2505.22375}, 
}

@article{ahmed2025qwen,
  title={Qwen 2.5: A comprehensive review of the leading resource-efficient llm with potentioal to surpass all competitors},
  author={Ahmed, Imtiaz and Islam, Sadman and Datta, Partha Protim and Kabir, Imran and Chowdhury, Md Naseef Ur Rahman and Haque, Ahshanul},
  year={2025},
  publisher={TechRxiv}
}

@article{grattafiori2024llama,
  title={The llama 3 herd of models},
  author={Grattafiori, Aaron and Dubey, Abhimanyu and Jauhri, Abhinav and Pandey, Abhinav and Kadian, Abhishek and Al-Dahle, Ahmad and Letman, Aiesha and Mathur, Akhil and Schelten, Alan and Vaughan, Alex and others},
  journal={arXiv preprint arXiv:2407.21783},
  year={2024}
}

@article{liu2024deepseek,
  title={Deepseek-v2: A strong, economical, and efficient mixture-of-experts language model},
  author={Liu, Aixin and Feng, Bei and Wang, Bin and Wang, Bingxuan and Liu, Bo and Zhao, Chenggang and Dengr, Chengqi and Ruan, Chong and Dai, Damai and Guo, Daya and others},
  journal={arXiv preprint arXiv:2405.04434},
  year={2024}
}

@inproceedings{ji2025towards,
  title={Towards economical inference: Enabling deepseek’s multi-head latent attention in any transformer-based llms},
  author={Ji, Tao and Guo, Bin and Wu, Yuanbin and Guo, Qipeng and Shen, Lixing and Chen, Zhan and Qiu, Xipeng and Zhang, Qi and Gui, Tao},
  booktitle={Proceedings of the 63rd Annual Meeting of the Association for Computational Linguistics (Volume 1: Long Papers)},
  pages={33313--33328},
  year={2025}
}

@article{meng2025transmla,
  title={Transmla: Multi-head latent attention is all you need},
  author={Meng, Fanxu and Tang, Pingzhi and Tang, Xiaojuan and Yao, Zengwei and Sun, Xing and Zhang, Muhan},
  journal={arXiv preprint arXiv:2502.07864},
  year={2025}
}

@inproceedings{ainslie2023gqa,
  title={Gqa: Training generalized multi-query transformer models from multi-head checkpoints},
  author={Ainslie, Joshua and Lee-Thorp, James and De Jong, Michiel and Zemlyanskiy, Yury and Lebr{\'o}n, Federico and Sanghai, Sumit},
  booktitle={Proceedings of the 2023 Conference on Empirical Methods in Natural Language Processing},
  pages={4895--4901},
  year={2023}
}

@misc{yu2026swaaslidingwindowattention,
      title={SWAA: Sliding Window Attention Adaptation for Efficient and Quality Preserving Long Context Processing}, 
      author={Yijiong Yu and Jiale Liu and Qingyun Wu and Huazheng Wang and Ji Pei},
      year={2026},
      eprint={2512.10411},
      archivePrefix={arXiv},
      primaryClass={cs.CL},
      url={https://arxiv.org/abs/2512.10411}, 
}

@inproceedings{NEURIPS2024_d13a3eae,
 author = {Yang, Songlin and Wang, Bailin and Zhang, Yu and Shen, Yikang and Kim, Yoon},
 booktitle = {Advances in Neural Information Processing Systems},
 doi = {10.52202/079017-3668},
 editor = {A. Globerson and L. Mackey and D. Belgrave and A. Fan and U. Paquet and J. Tomczak and C. Zhang},
 pages = {115491--115522},
 publisher = {Curran Associates, Inc.},
 title = {Parallelizing Linear Transformers with the Delta Rule over Sequence Length},
 url = {https://proceedings.neurips.cc/paper_files/paper/2024/file/d13a3eae72366e61dfdc7eea82eeb685-Paper-Conference.pdf},
 volume = {37},
 year = {2024}
}

@article{yang2024financial,
  title={Financial knowledge large language model},
  author={Yang, Cehao and Xu, Chengjin and Qi, Yiyan},
  journal={arXiv preprint arXiv:2407.00365},
  year={2024}
}

@article{merity2016pointer,
  title={Pointer sentinel mixture models},
  author={Merity, Stephen and Xiong, Caiming and Bradbury, James and Socher, Richard},
  journal={arXiv preprint arXiv:1609.07843},
  year={2016}
}

@inproceedings{wang2025alpha,
  title={Alpha-gpt: Human-ai interactive alpha mining for quantitative investment},
  author={Wang, Saizhuo and Yuan, Hang and Zhou, Leon and Ni, Lionel and Shum, Heung Yeung and Guo, Jian},
  booktitle={Proceedings of the 2025 Conference on Empirical Methods in Natural Language Processing: System Demonstrations},
  pages={196--206},
  year={2025}
}

@article{kim2024financial,
  title={Financial statement analysis with large language models},
  author={Kim, Alex and Muhn, Maximilian and Nikolaev, Valeri},
  journal={arXiv preprint arXiv:2407.17866},
  year={2024}
}

@article{yang2020finbert,
  title={Finbert: A pretrained language model for financial communications},
  author={Yang, Yi and Uy, Mark Christopher Siy and Huang, Allen},
  journal={arXiv preprint arXiv:2006.08097},
  year={2020}
}

@article{lu2023bbt,
  title={Bbt-fin: Comprehensive construction of chinese financial domain pre-trained language model, corpus and benchmark},
  author={Lu, Dakuan and Wu, Hengkui and Liang, Jiaqing and Xu, Yipei and He, Qianyu and Geng, Yipeng and Han, Mengkun and Xin, Yingsi and Xiao, Yanghua},
  journal={arXiv preprint arXiv:2302.09432},
  year={2023}
}

@article{wu2023bloomberggpt,
  title={Bloomberggpt: A large language model for finance},
  author={Wu, Shijie and Irsoy, Ozan and Lu, Steven and Dabravolski, Vadim and Dredze, Mark and Gehrmann, Sebastian and Kambadur, Prabhanjan and Rosenberg, David and Mann, Gideon},
  journal={arXiv preprint arXiv:2303.17564},
  year={2023}
}

@article{yang2023fingpt,
  title={Fingpt: Open-source financial large language models},
  author={Yang, Hongyang and Liu, Xiao-Yang and Wang, Christina Dan},
  journal={arXiv preprint arXiv:2306.06031},
  year={2023}
}

@misc{cmb2024yizhao,
  title={YiZhao-12B-Chat: A Chinese Financial Large Language Model},
  author={China Merchants Bank AI Lab},
  year={2024},
  publisher={ModelScope},
  url={https://www.modelscope.cn/models/CMB_AILab/YiZhao-12B-Chat/summary}
}

@article{zhu2025dianjin,
  title={Dianjin-r1: Evaluating and enhancing financial reasoning in large language models},
  author={Zhu, Jie and Chen, Qian and Dou, Huaixia and Li, Junhui and Guo, Lifan and Chen, Feng and Zhang, Chi},
  journal={arXiv preprint arXiv:2504.15716},
  year={2025}
}

@inproceedings{devlin2019bert,
  title={Bert: Pre-training of deep bidirectional transformers for language understanding},
  author={Devlin, Jacob and Chang, Ming-Wei and Lee, Kenton and Toutanova, Kristina},
  booktitle={Proceedings of the 2019 conference of the North American chapter of the association for computational linguistics: human language technologies, volume 1 (long and short papers)},
  pages={4171--4186},
  year={2019}
}

@article{raffel2020exploring,
  title={Exploring the limits of transfer learning with a unified text-to-text transformer},
  author={Raffel, Colin and Shazeer, Noam and Roberts, Adam and Lee, Katherine and Narang, Sharan and Matena, Michael and Zhou, Yanqi and Li, Wei and Liu, Peter J},
  journal={Journal of machine learning research},
  volume={21},
  number={140},
  pages={1--67},
  year={2020}
}

@article{brown2020language,
  title={Language models are few-shot learners},
  author={Brown, Tom and Mann, Benjamin and Ryder, Nick and Subbiah, Melanie and Kaplan, Jared D and Dhariwal, Prafulla and Neelakantan, Arvind and Shyam, Pranav and Sastry, Girish and Askell, Amanda and others},
  journal={Advances in neural information processing systems},
  volume={33},
  pages={1877--1901},
  year={2020}
}

@article{hu2022lora,
  title={Lora: Low-rank adaptation of large language models.},
  author={Hu, Edward J and Shen, Yelong and Wallis, Phillip and Allen-Zhu, Zeyuan and Li, Yuanzhi and Wang, Shean and Wang, Liang and Chen, Weizhu and others},
  journal={Iclr},
  volume={1},
  number={2},
  pages={3},
  year={2022}
}

@article{rafailov2023direct,
  title={Direct preference optimization: Your language model is secretly a reward model},
  author={Rafailov, Rafael and Sharma, Archit and Mitchell, Eric and Manning, Christopher D and Ermon, Stefano and Finn, Chelsea},
  }

@article{xiaomi2025mimo,
  title={MiMo: Unlocking the Reasoning Potential of Language Model--From Pretraining to Posttraining},
  author={Xiaomi, LLM and Xia, Bingquan and Shen, Bowen and Zhu, Dawei and Zhang, Di and Wang, Gang and Zhang, Hailin and Liu, Huaqiu and Xiao, Jiebao and Dong, Jinhao and others},
  journal={arXiv preprint arXiv:2505.07608},
  year={2025}
}

@article{yin2024entropy,
  title={Entropy Law: The Story Behind Data Compression and LLM Performance},
  author={Yin, Mingjia and Wu, Chuhan and Wang, Yufei and Wang, Hao and Guo, Wei and Wang, Yasheng and Liu, Yong and Tang, Ruiming and Lian, Defu and Chen, Enhong},
  journal={arXiv preprint arXiv:2407.06645},
  year={2024}
}

@inproceedings{lu2024instag,
  title={\#INSTAG: Instruction Tagging for Analyzing Supervised Fine-Tuning of Large Language Models},
  author={Lu, Keming and Yuan, Hongyi and Yuan, Zheng and Lin, Runji and Lin, Junyang and Tan, Chuanqi and Zhou, Chang and Zhou, Jingren},
  booktitle={International Conference on Learning Representations (ICLR)},
  year={2024}
}

@inproceedings{liu2024deita,
  title={What Makes Good Data for Alignment? A Comprehensive Study of Automatic Data Selection in Instruction Tuning},
  author={Liu, Wei and Zeng, Weihao and He, Keqing and Jiang, Yong and He, Junxian},
  booktitle={International Conference on Learning Representations (ICLR)},
  year={2024}
}

@article{hui2025decif,
  title={DecIF: Improving Instruction-Following through Meta-Decomposition},
  author={Hui, Tingfeng and Zhu, Pengyu and Ping, Bowen and Tang, Ling and Dong, Guanting and Zhang, Yaqi and Su, Sen},
  journal={arXiv preprint arXiv:2505.13990},
  year={2025}
}

@inproceedings{ouyang2024treecut,
  title={TreeCut: A Synthetic Unanswerable Math Word Problem Dataset for LLM Hallucination Evaluation},
  author={Ouyang, Jialin},
  booktitle={Proceedings of the 63rd Annual Meeting of the Association for Computational Linguistics},
  year={2025}
}

@inproceedings{wang2023selfinstruct,
  title={Self-Instruct: Aligning Language Models with Self-Generated Instructions},
  author={Wang, Yizhong and Kordi, Yeganeh and Mishra, Swaroop and Liu, Alisa and Smith, Noah A. and Khashabi, Daniel and Hajishirzi, Hannaneh},
  booktitle={Proceedings of the 61st Annual Meeting of the Association for Computational Linguistics},
  year={2023}
}

@article{xu2024wizardlm,
  title={WizardLM: Empowering Large Pre-trained Language Models to Follow Complex Instructions},
  author={Xu, Can and Sun, Qingfeng and Zheng, Kai and Geng, Xiubo and Zhao, Pu and Feng, Jiazhan and Tao, Chongyang and Lin, Qingwei and Jiang, Daxin},
  journal={arXiv preprint arXiv:2304.12244},
  year={2025}
}

\end{document}